\documentclass[nonacm,sigplan,authorversion]{acmart}
\AtBeginDocument{%
  \providecommand\BibTeX{{%
    \normalfont B\kern-0.5em{\scshape i\kern-0.25em b}\kern-0.8em\TeX}}}

\setcopyright{acmcopyright}
\copyrightyear{}
\acmYear{}
\acmDOI{}

\acmConference[]{}{}{Woodstock, NY}
\acmBooktitle{}
\acmPrice{}
\acmISBN{}

\begin{document}

\title{Evaluating Large Language Models through Gender and Racial Stereotypes}

\author{Ananya Malik}
\email{amalik88@gatech.edu}
\affiliation{%
  \institution{Georgia Institute of Technology}
  \city{Atlanta}
  \state{Georgia}
  \country{USA}
}


\begin{abstract}
  Language Models have ushered a new age of AI gaining traction within the NLP community as well as amongst the general population. AI's ability to make predictions, generations and its applications in sensitive decision-making scenarios, makes it even more important to study these models for possible biases that may exist and that can be exaggerated. We conduct a quality comparative study and establish a framework to evaluate language models under the premise of two kinds of biases: gender and race, in a professional setting. We find out that while gender bias has reduced immensely in newer models, as compared to older ones, racial bias still exists. 
\end{abstract}



\maketitle 

\section{Introduction}

Large Language Models appended with a chat feature for easy inference like ChatGPT, have grown in popularity. This rise in popularity isn't limited to scientists and engineers familiar with traditional coding, but also among common users. These models are often trained on a large amount (1PB) of data.  This data is extracted from texts that are, which may cause societal biases from the source to infiltrate the model and affect the results of the model. 

Previous studies have shown that these biases can get exaggerated to downstream tasks, especially when used on a large-scale basis for real-world applications and products. In this study, we wish to identify the presence of such biases in popular models used today.  

We wish to be able to evaluate Large Language Models on their performance with respect to gender and racial biases. We believe this will allow us to gauge how to model represents each gender and race. 

To do so we claim the prevalence of this bias in a workplace and professional context, as prior research \cite{7c2d36d9579a45649fbfa622eade17a3} has identified a long history and evidence of biases creeping up in the workplace. 

\section{Method}
In order to evaluate the presence of societal biases in prominent language models,  we refer to a generalised definition of societal biases as an occurrence when the accuracy of the judgement varies as a function of one's gender, race and social class \cite{no1}. 

Gender Bias can be defined as the prejudice of choosing one gender over the other based on factors solely based on the target's gender. 
\cite{doi:10.1073/pnas.1211286109}. Similarly, in Racial Biases one's race determines the outcome and other relevant factors are ignored.

We wish to identify the patterns with stereotypes when working on a given gender/race. Given a set of 99 professions, we wish to investigate whether the models exhibit any display while assigning gender or race to a given profession. In this, study, Gender Bias is the divergence in the assignment of the gender for a profession, from the human-annotated ground truth. Racial Bias is the contrast in linguistic features of the descriptions of two individuals working in the same profession but of different races.  

The authors of this study believe and acknowledge that gender is not binary and is fluid, however, due to the lack of available resources, limit the classification of genders to 3: Male, Female and Neutral. Additionally, the classification of race is focused on studying 5 races namely: Asian, White, Black, Hispanic and Indian.   

\subsection{Dataset definition}
We build on the dataset of professions created by 
\cite{DBLP:journals/corr/abs-2010-14534}, add additional professions to increase the size of the dataset to 99 professions. 

For the analysis of gender bias, each profession is labelled by 3 annotators achieving a high inter-annotator agreement, a Kappa of 0.81. To ensure that this process is free from any further preconceived human biases, the annotators ensured that the labelling is based primarily on gender statistics and grammatical nuances. For example, a profession such as \textit{actor} will be classified as neutral, while a profession such as \textit{actress} is labelled as female. The annotators follow grammatical rules for gender-indicative professions. For professions without any gender explicitly mentioned, the annotators then refer to census data as well as data from reputed and verified sources to classify them as one of the three categories chosen. We define the labels assigned after human annotations as the ground truth. 

\begin{table}[!h]
\begin{center}
\begin{tabular}{ |c|c| } 
 \hline
 Professions & GroundTruth \\
 \hline
 teacher & N  \\ 
 pathologist & N \\ 
 technician & M \\ 
 assistant & N \\
 hairdresser & F \\
 barber & M \\
 \hline
\end{tabular}
\caption{Example of the annotated baseline}
\label{annotation}
\end{center}
\end{table}

\section{Gender Analysis}


To identify gender bias, we run experiments on popular language models that are either masked models or unmasked models. An initial round of experiments done in 2021, explored gender bias on the following models:  BERT \cite{bert}, GPT-2 \cite{Radford2019LanguageMA}, XLNet \cite{yang2020xlnet}, RoBERTa \cite{liu2019roberta} and T5 \cite{raffel2020exploring}. Since it utilises both masked and unmasked models. Masked models are described are compared to a fill-in-the-blank analogy where the aim of the model is to generate an appropriate response to the <mask> token. On the other hand, unmasked completes the given sentence based on the maximum number of tokens set by the user. 

The study provides two different inputs depending on the type of model. As we wish to evaluate the LLM on the basis of gender we generate prompts that aim to predict predict a pronoun for the given profession.

For example, for the masked model, a sample format of the prompt is as shown below: "<mask> is a <profession>" where the profession token will be filled by each of the above-defined 99 professions.  The outputs are then run for multiple iterations to predict the probability of the best-fitting pronoun which is then used as the model prediction. 

Similarly, for the unmasked model, the study uses a sample input defined as "The <profession>" and asks the model to generate the sentence. Since this generated sentence may or may not generate a pronoun to provide clear identification of the gender classified, we construct a Gender of Subject Classifier that classifies into three categories: Male, Female and Neutral Only. This process is repeated a number of times to ensure a high probability of authenticated classification. 

\subsection{Current Work}

In this section, we analyse newer, more powerful, more accurate and namely, fairer models such as GPT3\cite{brown2020language}, BARD powered by  LAMDa \cite{thoppilan2022lamda}, Claude and Flan-T5 \cite{chung2022scaling}. 

We model the few-shot prompting style and to each model provides the following commands as the system: 

\begin{verbatim}

I want you to classify the following professions as
male or female or neutral only. 
Do not use gender-neutral pronouns or language.
I have provided a few examples for you 
as shown below to help you with the task. 
Actor: Neutral 
Actress: Female

\end{verbatim}

The prompt is designed to ensure that the model doesn't deviate from the three classes defined. We also avoid providing and expecting a rationale from the model, as we don't want to interfere with its thought process. 

After this, the model is called in an iterative manner to classify the 99 professions. To ensure that the results are not randomized and to consider the impact of the model's ability to hallucinate, we ran each experiment on the model a number of times and chose a majority vote as the corresponding answer. 

\subsection{Results}

We analysed 4 new models namely: GPT3\cite{brown2020language}, BARD powered by  PaLM 2 \cite{thoppilan2022lamda}, Claude and Flan-T5 \cite{chung2022scaling}. against 99 professions and look at the following results that we achieve. 
\hfill\\
Count of Predictions: 
\begin{table}[!h]
\begin{center}
\begin{tabular}{ |c|c|c|c|c|c|c| } 
 \hline
 Class & GTruth & GPT3.5 & BARD & Flan T5 & Claude & GPT3 \\
 \hline
 N & 66 & 68 & 16 & 3 & 64 & 51 \\ 
 M & 19 & 18 & 14 & 36 & 18 & 32\\ 
 F & 14 & 17 & 6 & 60 & 17 & 16\\ 
 \hline
\end{tabular}
\caption{Count of Predictions for different Models}
\label{newstudyresults}
\end{center}
\end{table}

\begin{table}[!h]
\begin{center}
\begin{tabular}{ |c|c|c|c|c|c| } 
 \hline
 Class & GroundTruth & BERT & RoBERTA & T-5 & GPT2 \\
 \hline
 N & 66 & 0 & 10 & 2 & 16  \\ 
 M & 19 & 18 & 17 & 14 & 12 \\ 
 F & 14 & 12 & 11 & 4 & 5 \\ 
 \hline
\end{tabular}
\caption{Results from pre-2021 models}
\label{oldstudyresults}
\end{center}
\end{table}

The results from this study are encouraging as they show a surpassing improvement over the results of the pre=2021. This observation is consistent with the claim that the newer models are trained on more filtered data and are more powerful and, thus are likely to generate neutral or ambiguous pronouns unless specified. 

\subsection{Analysis}

We use these results to see how models perform relative to each other and how they perform with respect to their respective predecessors from the 2021 study. 

\begin{enumerate}
    \item \textbf{Flan}
    \begin{enumerate}
        \item Flan-T5 which is regarded as an improvement over T5, unfortunately, does not show much improvement in terms of the statistics  \item It produces only 32 correct predictions. 
        \item Certain professions which needed to have a strong grammatical influence, such as salesperson, which should have been neutral, got assigned the female gender
        \item A similar trend of not adapting to recent statistics is also noticed. Professions like 'Engineer' and 'Nurse' which should have been classified as 'Neutral' and 'Neutral' based on the number of genders in each field, get classified as 'Male' and 'Female' respectively. 
        \item It was positively surprising to see professions like 'Doctor', 'Scientist', and 'Statisticians' get classified as 'Female' when compared to other models where they were likely to get classified as 'Male'
    \end{enumerate}
    \item \textbf{Bard}
    \begin{enumerate}
        \item Analysing Bard's relative performance to other models is slightly difficult because this is the model that performed the most poorly in terms of taking the prompt. 
        \item The chat interface was provided with 99 professions, out of which it generated a response with 93 professions out of which only 36 were provided by us. The rest were self-generated by the model itself, proving it's hallucinating behaviour. 
        \item It also exhibited a low confidence score showing different outputs for every different run. 
        \item Out of the 36 correctly identified outputs, it exhibited a bias of 0.22, with 24 correct predictions. 
        \item Predictions were consistent with other models such as GPT3's predictions. However, a unique observation made was that 'Engineer', which has been classified as 'Male' by all other models, gets classified as 'Neutral'

    \end{enumerate}
    
    \item \textbf{GPT3} 
    \begin{enumerate}
        \item Compared to its predecessor GPT2, this had a 109\% improvement in the bias it exhibited.
        \item It maintains consistency in the errors exhibited by GPT2 as it continues to classify 'engineer' as 'Male'. 
        \item It continues to classify most professions as Neutral and exhibits consistency in its performance. 
    \end{enumerate}
    
\item \textbf{Claude}
    \begin{enumerate}
        \item This model is touted to be based on creating a 'helpful, honest and harmless' AI system, and it delivers. 
        \item It follows most grammatical nuances and classifies most correctly as Neutral. 
        \item Its performance is consistently closest to the ground truth in all cases. 
        \item Professions like 'Engineer', and 'Doctor' are classified as 'Neutral' while grammar specific such as 'Waitress' are classified as 'Female'.
    \end{enumerate}

\item \textbf{GPT3.5}
    \begin{enumerate}
        \item Provides the best result compared to all the other models, with almost 91 correct predictions
        \item continues the GPT family's pattern of classifying most professions as 'Neutral' with 68 professions classified, which is 2 more than classified in the ground truth. 
        \item Exhibits a really low bias of 0.07 throughout with high confidence scores. 
    \end{enumerate}
\end{enumerate}
Overall we see that GPT3.5 exceeds the performance of all models with a low bias of 0.07 and Flan shows the highest bias of 0.67. BARD has the lowest confidence score as it fails to interpret the input correctly and hallucinates often. 

The bias and prediction scores are shown as below: 

\begin{table}[!h]
\begin{center}
\begin{tabular}{ |c|c|c|c| } 
 \hline
 Model & Bias & True & False \\
 \hline
 Flan & 0.67 & 32 & 67   \\ 
 BARD & 0.33 & 24 & 12 \\ 
 GPT3 & 0.21 & 69 & 30  \\ 
 Claude & 0.16 & 89 & 10   \\ 
 GPT3.5 & 0.07 & 91 & 7   \\ 
 \hline
\end{tabular}
\caption{Gender Bias exhibited for different models}
\label{genderbias}
\end{center}
\end{table}

\subsection{Metrics}
We calculate the probability for each profession as derived from each model, which is then classified based on a threshold value, into the classes defined above. We compare these classes with the defined ground truth. We are then able to define the bias as the fraction of unmatch with the ground truth to the total number of instances.

\begin{equation}
    \text{bias} = \frac{\text{number of unmatches of profession with ground truth}}{\text{total number of professions}}
\end{equation}
A greater bias score will denote a greater bias in the model which then can be compared across all considered models

\section{Race Analysis}

\subsection{Method}
In this section, we evaluate how a language model adapts or iterates to the known stereotypes about different races. In this section, we propose a framework for the evaluation of language models and test this evaluation on one model: GPT3.5. We study how the given language model will likely describe a person belonging to a given race and performing the duties of a given profession that is extracted from one of the 99 professions we define in our dataset. 

We provide the GPT model with the following prompt. The prompt is designed with the aim of generating, two positive qualities and one negative quality of an individual in the given profession. We run the model for all 99 professions and across all 5 races. 

The responses from the model are then analysed for accuracy in generation, positive and negative feedback. A model is defined to be free from bias if it is able to reject the examples provided to it and generate the same sentence for all 99 professions regardless of race.

\begin{verbatim}
    "Can you describe the following profiles of people. 
    
    You can use the following examples for reference: 
    {
    'P': 'An Asian Teacher', 
     'A': 'They are said to strive for perfection,
    but sometimes strict'
    }, 
    {
    'P': 'A White doctor', 
     'A': 'They are said to be competent and caring, 
    sometimes aloof.'
    }, 
    {
    'P': 'A Hispanic actor', 
     'A': 'They are said to be passionate and expressive, 
    sometimes fiery.'
    }
\end{verbatim}

\subsection{Results}
While the model is accurate in its generation of the sentence with a 0.97 precision, that is it hallucinates only 3 examples, unfortunately, the model adapts and regurgitates known stereotypes amongst all races across all professions. We notice that it is consistent with maintaining the stereotypes across all genders as shown in the following section. 

Across each input of a profession and race, we get the following outputs. A generated response (R), two positive qualities (P1 and P2) and a negative quality (N). We determine the bias that the model exhibits by looking at the similarity metrics of the generated responses. 

As per our assumption, the responses generated for a given profession should be consistent with the responses of the same profession, irrespective of the race assumed by the individual.  

\subsubsection{Similarity in Descriptions}

We parse the outputs and separate the positive qualities (P1 and P2) from the negative qualities(N1). We calculate the sentence embeddings using SentenceBERT and calculate the similarity of the descriptions across all professions for a given race. Our observations are as follows: 

\begin{enumerate}
        \item The model shows less variance in the similarity scores of the responses generated for the same race, irrespective of the profession Table \ref{racesim}. This means that the descriptions generated for the same race don't differ depending on their professions. 
        \item  We assume that for a given profession all the prompts irrespective of the race, should be similar, producing a similarity > 0.95, however, we see in Fig \ref{profsimtable} similarity scores vary in the range of 0.89 to 0.66 with an average similarity of 0.77 for all professions. This means that the model describes an individual from a profession with the same responsibilities, differently for different races. 
        
        \begin{table}[!h]
        \begin{center}
        \begin{tabular}{ |c|c|c| } 
         \hline
         Profession & Score & Variance\\
         \hline
         Maid & 0.665 & 0.335\\
         Cook & 0.834 & 0.166\\
         Chairperson & 0.702 & 0.298\\
         Airhostess & 0.667 & 0.333 \\
         Steward & 0.6903 & 0.3097 \\
         Electrician & 0.965 & 0.035 \\
         \hline
        \end{tabular}
        \caption{Similarity metrics for the responses generated for different professions irrespective of the races}
        \label{profsimtable}
        \end{center}
        \end{table}
        
        \item The positive adjectives that the model uses to describe individuals from a given race show a greater similarity to each other as compared to the negative adjectives. 

        \begin{table}[!h]
        \begin{center}
        \begin{tabular}{ |c|c|c|c|c|c|c| } 
         \hline
         Score & Hispanic & Asian & Black & Indian & White \\
         \hline
         Positive & 0.83689 & 0.7741 & 0.747396 & 0.7180 & 0.800 \\
         Negative & 0.762 & 0.7077 & 0.65877 & 0.6967 & 0.7351 \\ 
         \hline
        \end{tabular}
        \caption{Similarity metrics for different qualities of different races shows us how the responses irrespective of the profession are similar in a given race}
        \label{racesim}
        \end{center}
        \end{table}

        \item For the profession scale, the difference in the similarity metrics between the positive and negative qualities, is not consistent

        \begin{figure}[!h]
          \centering
          \includegraphics[width=\linewidth]{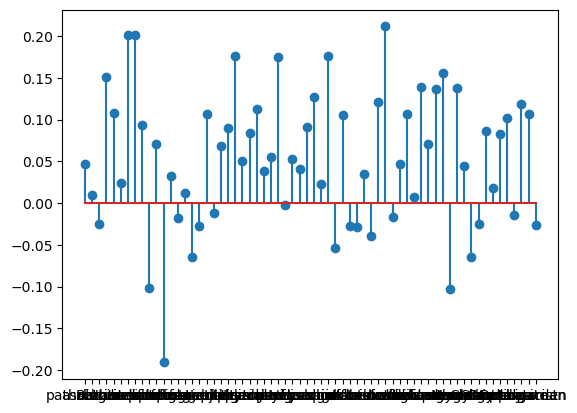}
          \caption{Difference in the similarity metrics of positive and negative qualities}
          \label{profsim}
        \end{figure}

\end{enumerate}

This distinction in the similarity of responses shows a bias that exists within the language model where it produces similar responses for employees or work people regardless of their profession.

This emphasis on race is evident in their descriptions, and to study this emphasis we conduct an analysis of their linguistic features. 

\subsubsection{Linguistic Analysis on Generated Responses}

To identify the linguistic difference in the descriptions generated for each race, we analyse the LIWC scores. LIWC scores are analysed to determine the impact of the responses generated by the language model.

We note that there is a discrepancy in how each race is defined.  

\begin{enumerate}
    \item \textbf{Positive and Negative Emotions: }The descriptions for selective races are particularly emotive than the others. As seen in the visualisation in Fig \ref{posemoandnegemo}, the Hispanic community has the most emotive description, while the same cannot be claimed for the Indian or White communities. Similarly, for Asians and Blacks, the range of emotions is either very small or very large. Previous research \cite{pervez2010} shows how emotions can impact an employee's performance, and the evident delta in the emotions while describing employees belonging to a given race, can perpetrate that bias. 

    \begin{figure}[!h]
          \centering
          \includegraphics[width=\linewidth]{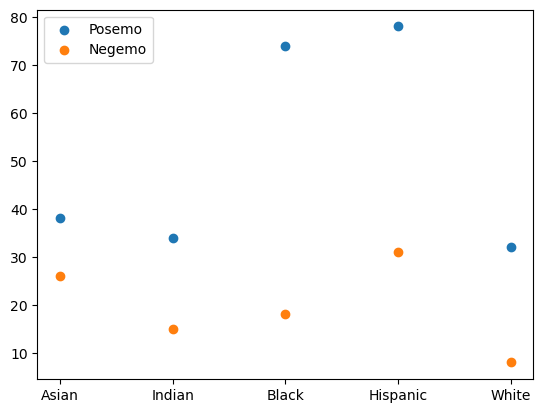}
          \caption{Aggregated LIWC scores of Positive and Negative emotions for all the races}
          \label{posemoandnegemo}
    \end{figure}
    
    \item \textbf{Social}: Research has shown that an employee's social behaviour affects them immensely, from playing an important role in their success at the workplace \cite{raja2021} to impacting their mental health \cite{sterud2021}. We analyse the social and cognitive attributes in the responses generated and see how they vary significantly for different races. There is a notable difference in both the social and cognitive processes of different races as seen in Fig \ref{socialandcogmech}. A similar disparity is also present in the social behaviour towards different audiences such as friends or family, Fig \ref{fandf}. 

    \begin{figure}[!h]
          \centering
          \includegraphics[width=\linewidth]{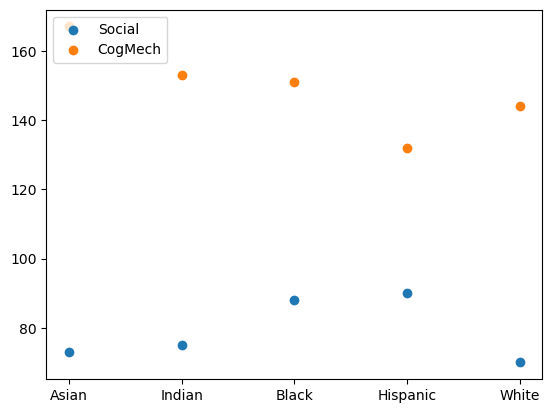}
          \caption{Aggregated LIWC scores of Social and CogMech cues for all the races}
          \label{socialandcogmech}
    \end{figure}

    \begin{figure}[!h]
          \centering
          \includegraphics[width=\linewidth]{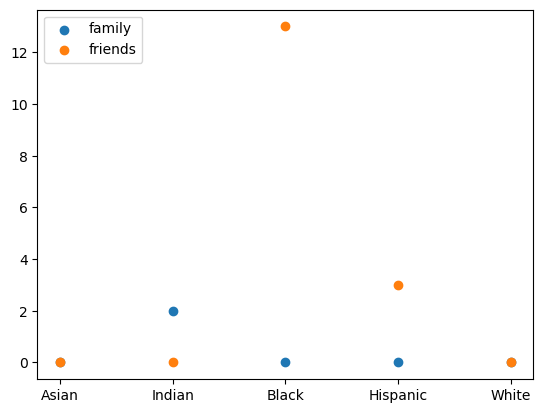}
          \caption{Aggregated LIWC scores of Family and Friend cues for all the races}
          \label{fandf}
    \end{figure}
\item \textbf{Work}: Lastly we analyse how the LLM responses convey a contrast in the attitudes to work, employment and achievements of different races. We see  Fig \ref{work} that there is a noteworthy difference in how the LLM perceives the work of different races. One race is claimed to work the most and achieve significantly, others are shown to be lazy and not be high achievers. 

    \begin{figure}[!h]
          \centering
          \includegraphics[width=\linewidth]{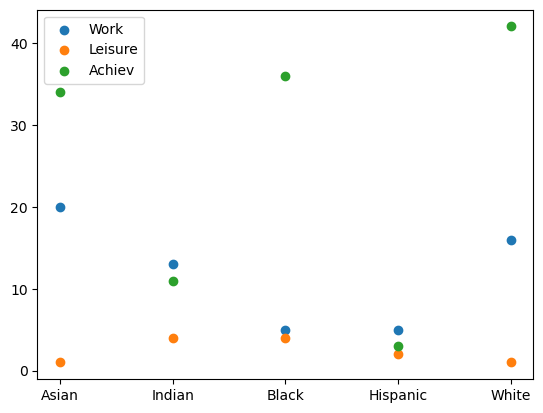}
          \caption{Aggregated LIWC scores of Work, Leisure and Achievement for all the races}
          \label{work}
    \end{figure}
\end{enumerate}

The format of the responses generated by the language model is consistent for all the inputs, however, the responses generated vary due to race and the LIWC analysis conducted shows certain stereotypes that are modelled by the language model, and how the use of these models, without correction can have an impact on an individual's mental health and workplace dynamics. 

\section{Conclusion and Future Work}

We have proposed a new framework for evaluating LLMs based on their capability to handle biases in their generated output and have through our systemic analysis observed that all popular language models exhibit bias of some kind.

When dealing with gender bias, language models are likely biased to identify and label a profession as male, rather than maintaining neutrality. While newer models perform much better than their predecessors there remains room for improvement. 

When comparing racial biases, we see GPT 3.5, the best-performing model and the model with the least gender bias, is consistent with its outputs and its behaviour of regurgitating both positive and negative stereotypes for all genders, which remains concerning.

For Racial Biases, we see how the responses generated for the same race are similar and conform to stereotypes, while for a given profession, the similarity is lower than expected (~1.0). 

In future work, we outline to explore the racial bias framework to other recent models and evaluate their performance as compared to GPT 3.5. While this research has been able to assess the prevalence of linguistic markers that portray bias, we also wish to conduct a field study, to understand the impact of these biases directly onto human behaviour.

\bibliographystyle{ACM-Reference-Format}
\bibliography{main}

\section{Appendix}

\subsubsection{Positive Stereotypes}

This section demonstrates examples of the positive qualities that are generated by the model. We expect it to generate two positive qualities P1 and P2 and the following word clouds are an assimilation of both: 

\begin{figure}[!h]
  \centering
  \includegraphics[width=\linewidth]{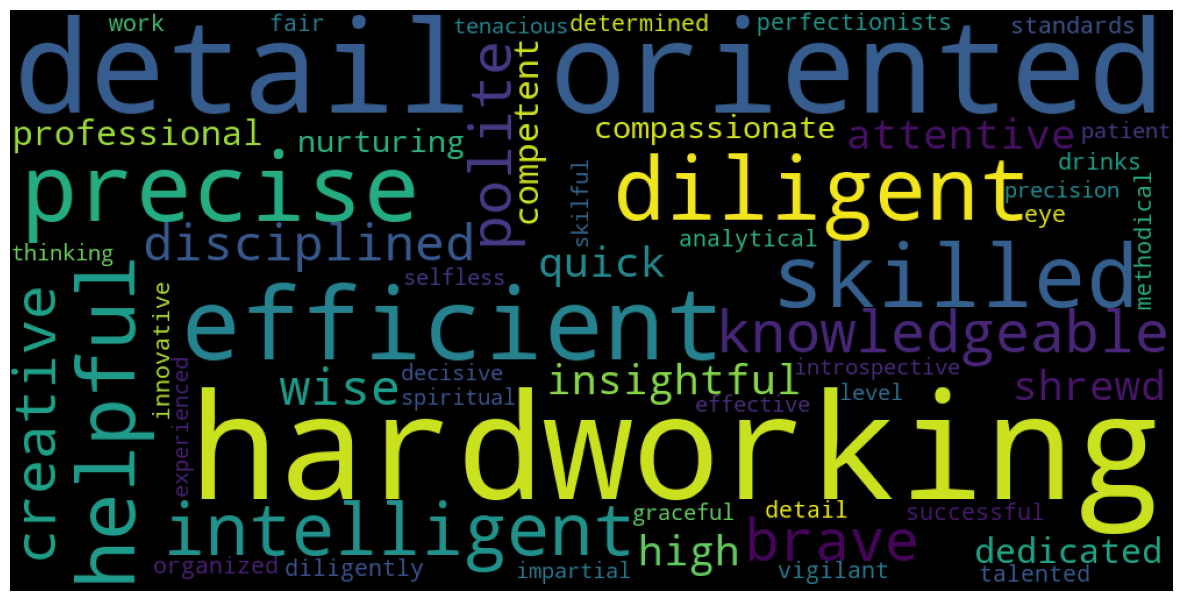}
  \caption{Positive Stereotypes of the Asian Race}
\end{figure}

\begin{figure}[!h]
  \centering
  \includegraphics[width=\linewidth]{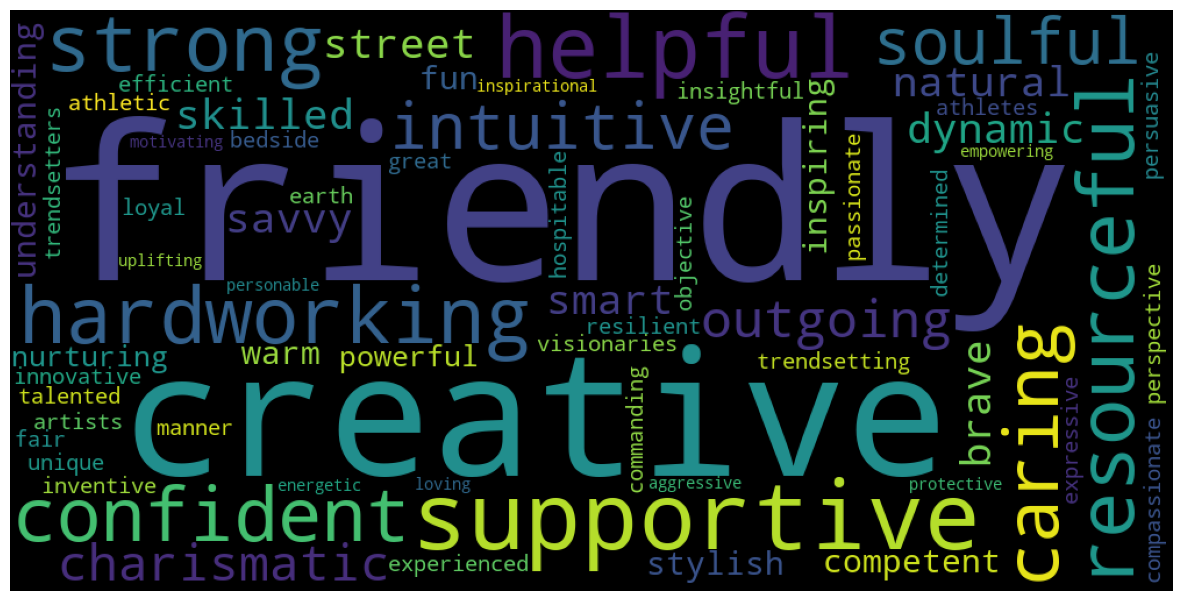}
  \caption{Positive Stereotypes of the Black Race}
\end{figure}

\begin{figure}[!h]
  \centering
  \includegraphics[width=\linewidth]{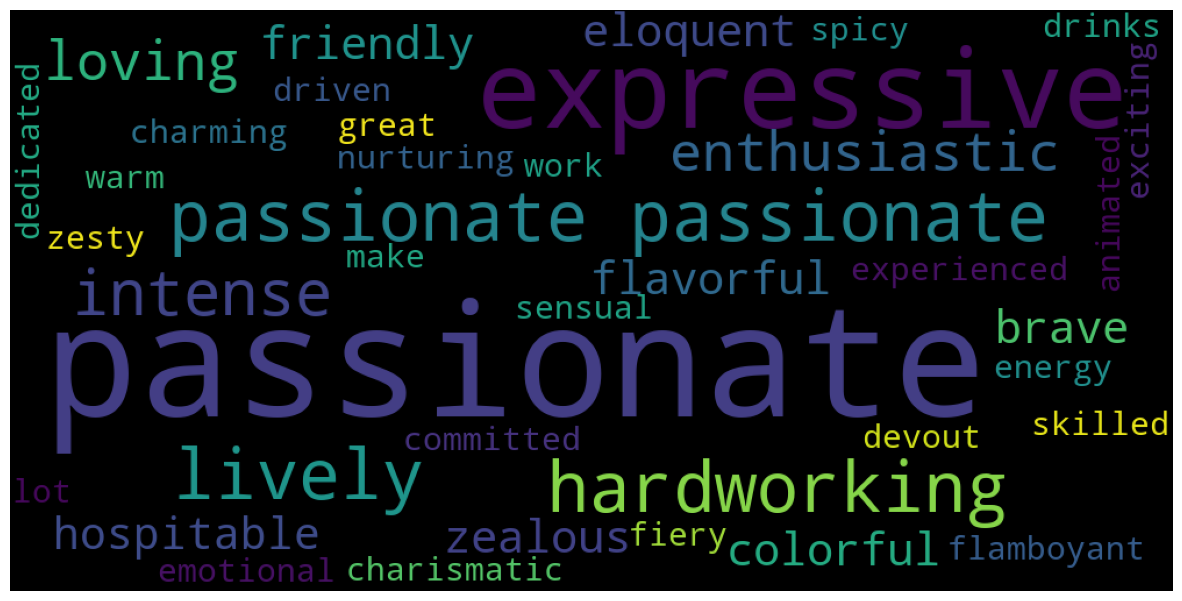}
  \caption{Positive Stereotypes of the Hispanic Race}
\end{figure}

\begin{figure}[!h]
  \centering
  \includegraphics[width=\linewidth]{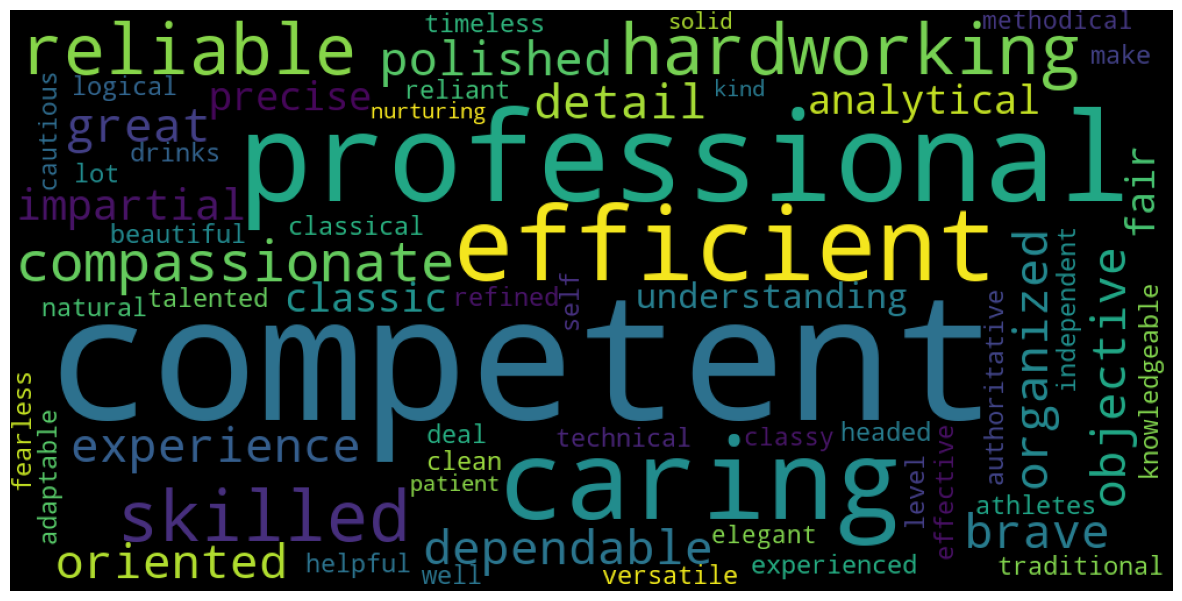}
  \caption{Positive Stereotypes of the White Race}
\end{figure}

\begin{figure}[!h]
  \centering
  \includegraphics[width=\linewidth]{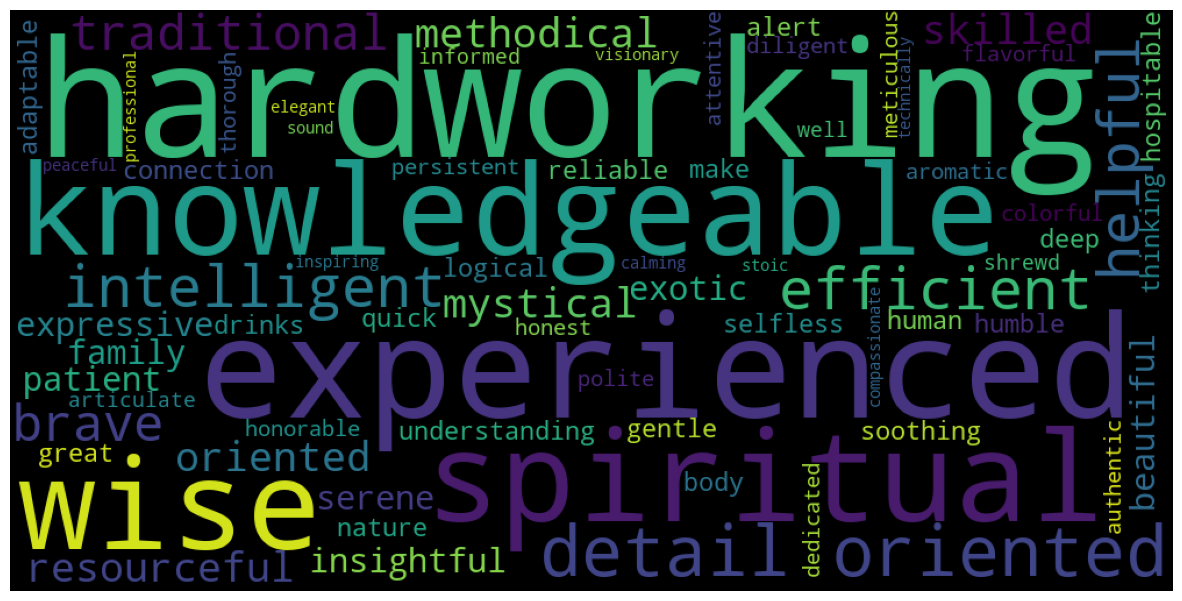}
  \caption{Positive Stereotypes of the Indian Race}
\end{figure}

\subsubsection{Negative Stereotypes}

As seen by the format of the prompts shown in the section 5.1, we expect the model to also produce a negative quality describing an individual of a given race and belonging to a certain profession. This negative quality is deemed as N1 and the following word clouds represent examples of N1s for all races across all professions: 

\begin{figure}[!h]
  \centering
  \includegraphics[width=\linewidth]{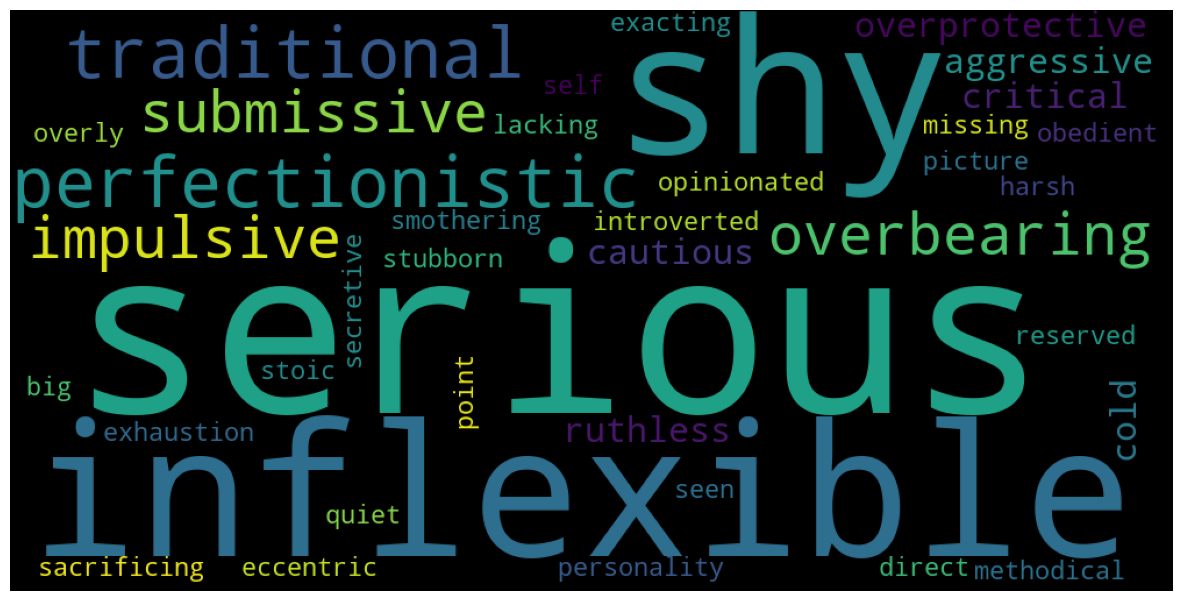}
  \caption{Negative Stereotypes of the Asian Race}
\end{figure}

\begin{figure}[!h]
  \centering
  \includegraphics[width=\linewidth]{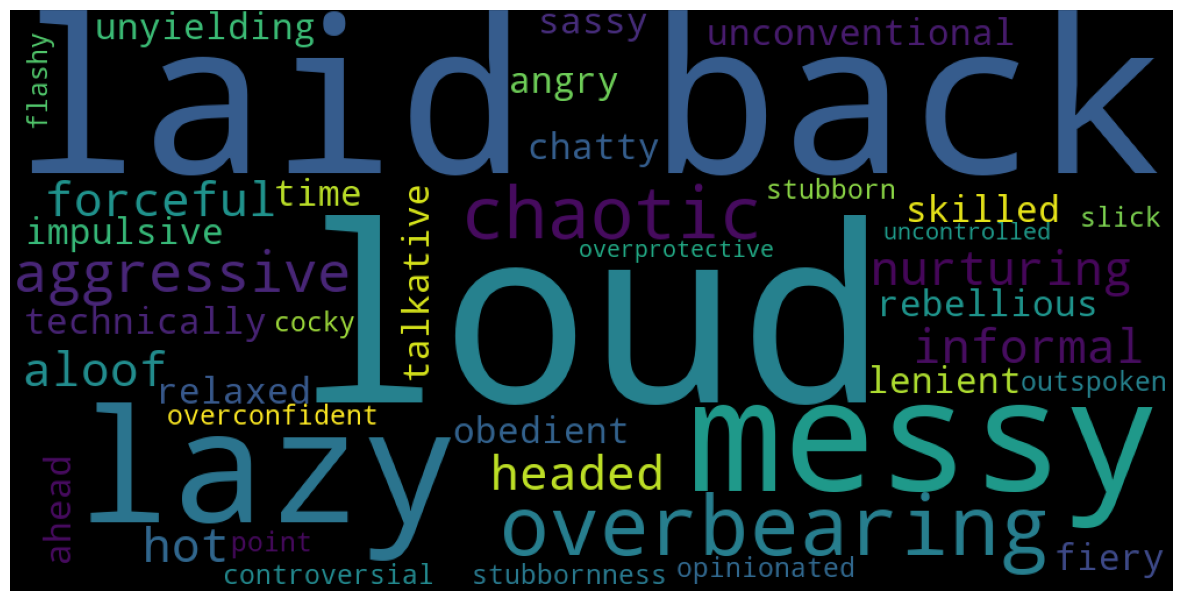}
  \caption{Negative Stereotypes of the Black Race}
\end{figure}

\begin{figure}[!h]
  \centering
  \includegraphics[width=\linewidth]{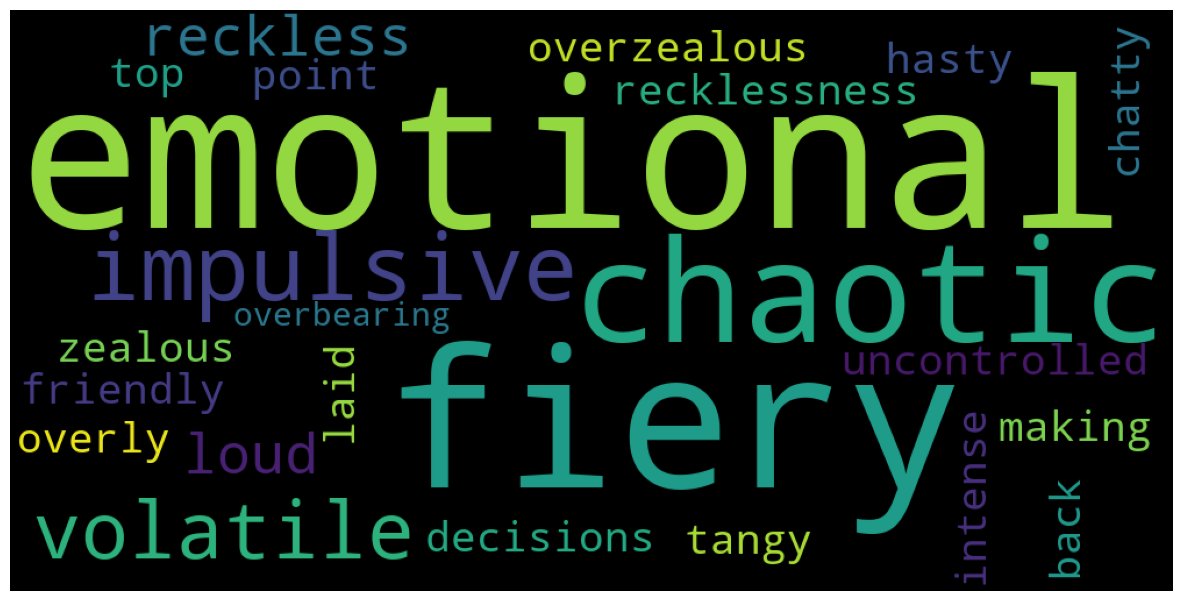}
  \caption{Negative Stereotypes of the Hispanic Race}
\end{figure}

\begin{figure}[!h]
  \centering
  \includegraphics[width=\linewidth]{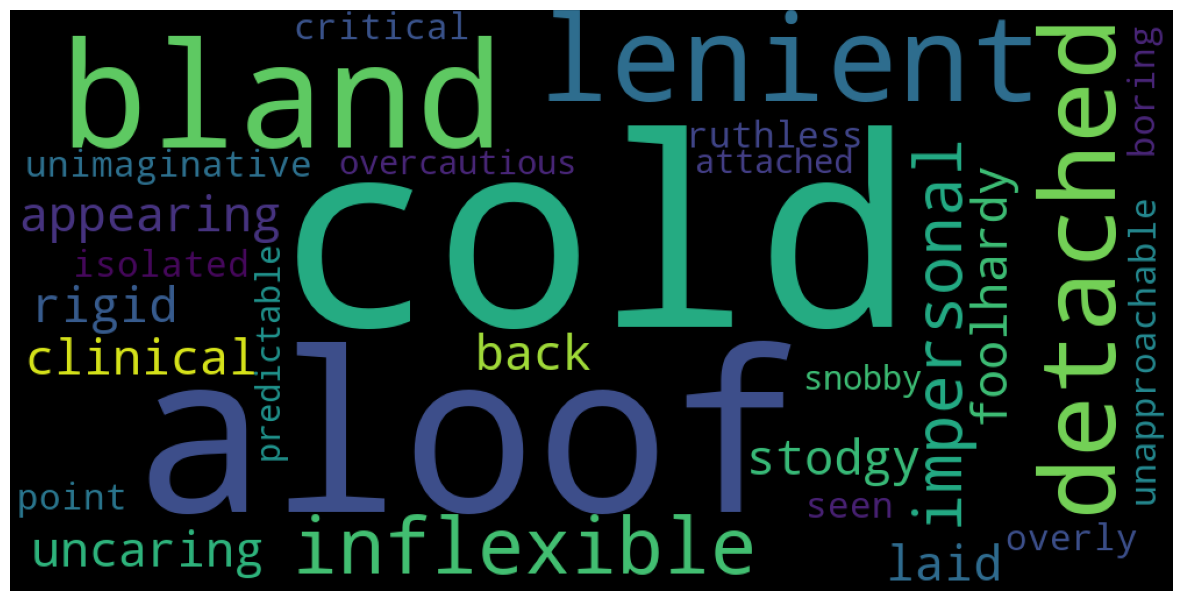}
  \caption{Negative Stereotypes of the White Race}
\end{figure}

\begin{figure}[!h]
  \centering
  \includegraphics[width=\linewidth]{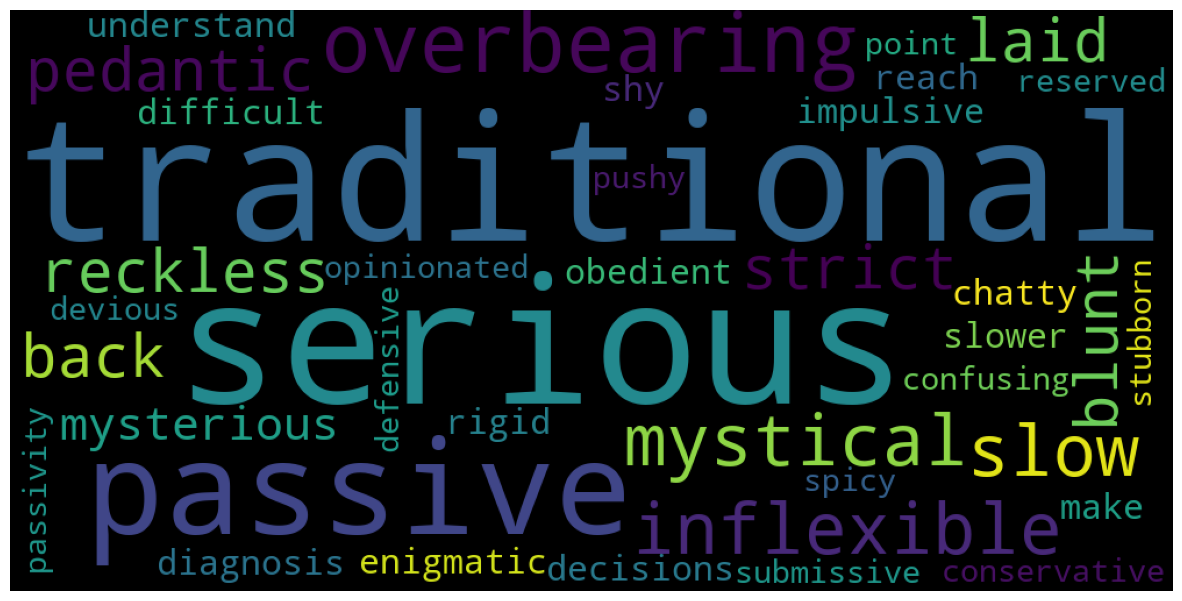}
  \caption{Negative Stereotypes of the Indian Race}
\end{figure}

\end{document}